# $L_0$ Regularization of Field-Aware Factorization Machine through Ising Model

Author: Yasuharu Okamoto[1,2]

[1]Secure System Platform Research Laboratories, NEC Corporation, 1753 Shimonumabe, Nakahara-ku, Kawasaki, Kanagawa 211-8666, Japan

[2]NEC-AIST Quantum Technology Cooperative Research Laboratories, 1-1-1 Umezono, Tsukuba, Ibaraki, 305-8568, Japan

Corresponding author: E-mail: y-okamoto@aist.go.jp

## Abstract

We examined the use of the Ising model as an $L_0$ regularization method for field-aware factorization machines (FFM). This approach improves generalization performance and has the advantage of simultaneously determining the best feature combinations for each of several groups. We can deepen the interpretation and understanding of the model from the similarities and differences in the features selected in each group.

## Introduction

The Ising model approach to solving combinatorial optimization problems consisting of 0/1 binary variables has recently attracted significant attention.[1-8] This is partly due to the promise of analog quantum computation such as quantum annealing[9] and cold atoms.[10] However, for now, quantum-inspired classical computers[11] and algorithms[12-14] will play a significant role in the number of available bits. Since the approach is heuristic, it only sometimes yields exact solutions, and dedicated solvers are more efficient for specific problems that have been well-studied for a long time, such as TSP.[15] However, the Ising model can be applied to various combinatorial optimization problems and helps solve industrial and social problems.[1-8]

It is noteworthy that constructing the Ising model from experimental/computational data and other observations to find the optimal solution using factorization machine (FM)[16,17] is also gaining significant attention.[6] In this paper, we use a field-aware factorization machine (FFM),[18] an extension of FM.

In FM and FFM, there is a substantially increased number of features due to the existence of feature cross terms, and overfitting becomes a severe problem. A common countermeasure against overfitting in machine learning is to introduce $L_1$ regularization or $L_2$ regularization,[19] but it is challenging to integrate $L_1$ regularization with optimization through the Ising model. On the other hand, $L_0$ regularization is usually not as practical as $L_1$ and $L_2$ regularization because it is not differentiable, and the feature selection becomes a combinatorial problem, resulting in a significant computational load. However, the drawback is not a problem in the case of combinatorial optimization through the Ising model but rather a good match.

$L_0$ regularization builds a sparse learning model that is computationally efficient for predictions, making it suitable for edge computing. The advantage of $L_0$ regularization in directly controlling the number of features included in the model is also helpful for model interpretation. Besides, it is possible to control the total number of features to be selected and the number of features in a given group. The control of the number of features can avoid the multicollinearity problem seen with $L_1$ regularization, i.e., the problem of selecting only one arbitrary feature in a group of strongly correlated features.

FM and FFM can handle continuous variables, but it is complicated and inefficient to handle continuous variables with the Ising model, so here, we arrange the quantitative values in ascending order to make them categorical variables. There is a concern that this will reduce the descriptive power of the model due to the conversion from continuous variables to categorical variables, but this is different, as shown in the Results and Discussion section. Moreover, it allows the optimal features to be selected for each category simultaneously. For example, it is possible to classify age into four categories: young, middle-aged A, middle-aged B, and elderly people, and construct an optimal model for each with $L_0$ regularization simultaneously.

Since this study converted quantitative variables, the original features, into categorical variables, we set the FFM field to represent the original features. For example, the age field contains the above four categories (young, middle-aged A, middle-aged B, and elderly). Thus, features essential for the younger age group may differ from those critical for the older age group. The similarities and differences in the selected features among the groups contribute to the interpretation and understanding of the obtained model.

Such feature selection with 0/1 binary variables has commonality with selecting valid partial networks by taking the logical product of the weights of random initial values of hidden neural networks and super masks represented by 0/1 binary variables to construct lightweight neural networks.[20] The ability to select features for each group is also effective as a countermeasure against Simpson's paradox.[21]

In this paper, to specifically investigate the approach combining FFM and $L_0$ regularization with the Ising model, we used a dataset that examined the relationship between features such as age, gender, BMI, and blood pressure levels of diabetic patients and the progression of their disease after one year.[22] The proposed method showed generalization performance comparable to or better than the regression model with $L_1$ regularization with the original continuous variables.

## Computational Methods

**FFM:** In FM,[16,17] the predicted value $\hat{y}_i$ for the $i$-th data when considering up to the second-order cross term is given by,

$$\hat{y}_i = w_0 + \sum_{l=1}^{D} w_l x_{il} + \sum_{l_2 > l_1}^{D} w_{l_1 l_2} x_{il_1} x_{il_2} \quad (1)$$

Here, $x_{il}$ is a 0/1 binary variable for the $l$-th feature of the $i$-th data and is given from the input dataset. $D$ is the number of features. In addition, $w_0, w_l, w_{l_1 l_2}$ are parameters determined by FFM learning based on training data. In FFM,[18] the following approximation is used to express the second-order parameter ($w_{l_1 l_2}$) as the inner product of the latent vector $v$ expanded in $K$ terms (we used $K = 4$ throughout the paper).

$$w_{l_1 l_2} \sim \sum_{m=1}^{K} v_{l_1 f(l_2)}^m v_{l_2 f(l_1)}^m \quad (2)$$

Note that $f(l)$ refers to the field to which the $l$-th feature belongs. We optimized the parameters $w_0, w_l, w_{l_1 l_2}$ by applying the stochastic gradient descent (SGD) method[23] to the residual sum of squares in equation (1). The learning rate $\eta$ of the SGD is assumed to be the below formula (3).

$$\eta(n_e) = a_1 exp(-a_2 \times n_e) \quad (3)$$

Where $n_e$ corresponds to epoch, and the total number of epochs is set to 300. We used cross-validation with grid search to determine hyperparameters $(a_1, a_2)$.

**Categorical variables:** Here, we used a dataset that examined the relationship between 10 features such as age, gender, BMI, and blood pressure level of diabetic patients and the progression of the disease after one year. The data is standardly included in a free software machine learning library, scikit-learn (sklearn.datasets.load_diabetes).[22] The total number of data is 442.

Ising model optimization is easy to formulate when we employ 0/1 binary variables as explanatory variables. Therefore, in the examples described in the next section, an ordinal scale was introduced for the remaining nine quantitative variables, excluding gender (already a categorical variable), and we transformed each variable into a categorical variable consisting of four groups. Specifically, each quantitative variable was ordered ascendingly and divided into four groups by quartiles. Here, grouping by quartiles was employed to avoid significant variations in the number of data belonging to each group. As stated, each field corresponds to each feature before converting it into a categorical variable, so the number of fields is 10.

The nine features other than gender, initially categorical variables, were divided into four groups, resulting in 38 features (only gender has two groups, and the others have four groups). These features generate 703 cross terms; thus, 741 (= 38 + 703) variables for the Ising model optimization will be referred to as expanded features and explained in the following subsection. Furthermore, as will be explained later, since we focus on a specific field and determine the learning model concerning four groups in the field, respectively, the size of QUBO (quadratic unconstrained binary optimization) matrix is 2964 × 2961 (= [4 × 741] × [4 × 741]). The FFM learning parameters are 1, 38, and 1520, respectively.

**Expanded features:** We define the subscript $s$ below, which represents the subscripts of the linear ($w_l$) and quadratic ($w_{l_1 l_2}$) terms in equation (1) together in one dimension.

$$s = s \quad (1 \leq s \leq D), \quad (l_1, l_2) \to s \quad (D + 1 \leq s \leq D(D+1)/2)$$

The latter part can be written in detail as shown in Table 1. This subscript is used to collectively represent the features (first order) and their cross terms (second order), so it is called the expanded features.

Table 1: Relationship between one-dimensional subscript $s$ and two-dimensional subscript $(l_1, l_2)$

| $(l_1, l_2)$ | $(1, 2)$ | $(1, 3)$ | · · · · · · · · · · · · · · · · · | $(D - 1, D)$ |
|---|---|---|---|---|
| $s$ | $D + 1$ | $D + 2$ | · · · · · · · · · · · · · · · · · | $D(D + 1)/2$ |

**Feature selection through the Ising model:** Besides the binary variables $x_{il}$ defined above, we introduce binary variables $p_{ig}$: $p_{ig} = 1$ if the $i$-th data belongs to group $g$, which determined from the input data, $p_{ig} = 0$ otherwise. We can rewrite the residual $\mathcal{L}_i$ using the subscript $s$ to identify the expanded features,

$$\mathcal{L}_i \equiv y_i - \hat{y}_i = (y_i - w_0) - \sum_{s,g} \alpha_s X_{is} p_{ig} q_{sg} \quad (4)$$

Where $X_{is}$ and $\alpha_s$ are,

$$X_{is} = \begin{cases} x_{is} & (1 \leq s \leq D) \\ x_{il_1} x_{il_2} & (D + 1 \leq s \leq D(D+1)/2) \end{cases} \quad (5)$$

$$\alpha_s = \begin{cases} w_s & (1 \leq s \leq D) \\ w_{l_1} w_{l_2} & (D+1 \leq s \leq D(D+1)/2) \end{cases} \quad (6)$$

Here, $D(=38)$ is the number of features, $q_{sg}$ is a binary variable optimized through the Ising model, and if group $g$ selects the $s$-th expanded feature, $q_{sg} = 1$, otherwise $q_{sg} = 0$. For $L_0$ regularization, it is necessary to set the penalty function so that the sum of the expanded features selected in each group is $M_f$. Therefore, the objective function $\mathcal{F}$ for optimization through the Ising model is as follows.

$$\mathcal{F} = \sum_i \left( y_i - w_0 - \sum_{s,g} \alpha_s X_{is} p_{ig} q_{sg} \right)^2 + A \sum_g \left( \sum_s q_{sg} - M_f \right)^2 \quad (7)$$

The second term on the right side represents the penalty function. $A$ is a hyperparameter that determines the strength of the constraint term, and here, we set it to $A = 10$ (this satisfied the constraint condition in all calculation examples). In the discussion below, we evaluate the generalization performance by optimizing $q_{sg}$ using the FFM training data and applying the determined $q_{sg}$ to the test data. We used tabu search (TS)[24, 25] to optimize equation (7) because TS gave better results than simulated annealing in our preliminary tests.

# Results and Discussion

**Coefficient of determination of training/test data for diabetes dataset with FFM:** First, we examined the coefficient of determination ($R^2$) for training and test data when FFM was used alone without combining with $L_0$ regularization. The total number of data was 442, and we divided the data into 3 (training data) : 1 (test data). Furthermore, we divided the training data into five blocks. We used cross-validation and grid search to determine the learning rate hyperparameters $(a_1, a_2)$ included in equation (3) and found the optimal values $(a_1, a_2) = (1.0^{-3}\sqrt{10}, 1.0^{-2}\sqrt{10})$. Using the values, we created a model that uses all five blocks as training data and calculated $R^2$ for the test data, which was 0.4181 ($R^2$ for the training data was 0.5768).

In order to compare these values, we performed calculations on the original data without converting it into categorical variables, choosing random forest (RF) and elastic net (EN) as typical regression methods by using scikit-learn's APIs (GridSearchCV etc.).[23] The results of $R^2$ were 0.4122 (RF) and 0.4863 (EN) for the training data, and 0.4689 (RF) and 0.4728 (EN) for the test data. Note that the optimal value of EN (l1_ratio = 1.0) does not include the $L_2$ regularization term and degenerates to $L_1$ regularization.

$R^2$ value for the training data of the FFM exceeds that of both RF and EN. The number of features is substantially increased by categorization, enhancing the model's descriptive power.

On the other hand, the FFM result for the test data is lower than that of the existing methods. The difference between training and test data suggests overfitting and the importance of regularization.

**Relationship between the number of features selected for $L_0$ regularization and generalization performance:** The character of the present approach is that $L_0$ optimization with the Ising model is performed afterward on the model construction by FFM. Therefore, it is easy to investigate how the number of features selected for $L_0$ regularization affects generalization performance. In addition, we check the effect of converting the quantitative variables into categorical variables with quartiles. For example, the age field contains four groups: young, middle-aged A, middle-aged B, and elderly. By optimizing equation (7), we can determine each group's optimal set of features in equal numbers but different combinations. There are two cases where the features selected by the two groups are common to both groups, and the other is unique to each group. Comparing both cases may provide helpful information for understanding the dataset.

Figure 1 shows the change of $R^2$ for the training and test data when the number of expanded features selected by $L_0$ regularization ($M_f$) is varied. As mentioned above, although the calculation corresponds to dividing the age field into four groups by quartiles, this figure shows the results for all groups together rather than for each group (we will discuss individual groups in the next subsection). It is also easy to construct a learning model based on groups in other fields, such as BMI, instead of age. However, in our attempt, the results classified by age were the most interesting.

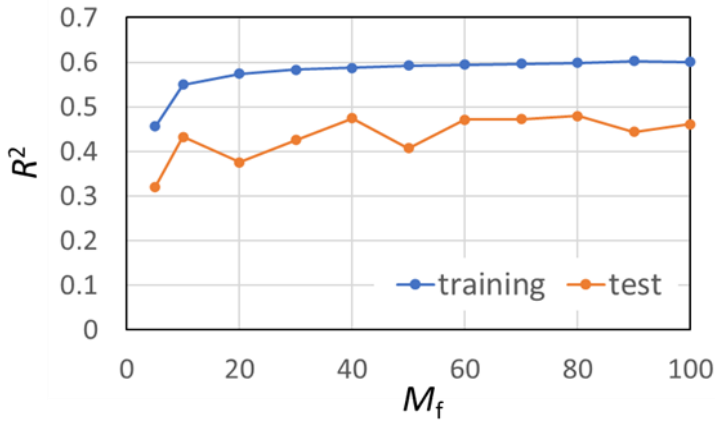

Figure 1: Relationship between the number of expanded feature selections ($M_f$) and the coefficient of determination ($R^2$) when all groups are combined.

For $M_f \geq 30$, $R^2$ for the test data tends to be larger than that for FFM alone (0.4181), as expected for the regularization effect. The result indicates an improvement in generalization performance. In particular, for $M_f = [40, 60, 70, 80]$, it is comparable to the values obtained with RF(0.4689) and EN (0.4728). Thus, converting to categorical variables was fine.

Similarly, the introduction of $L_0$ regularization did not pose any significant problems. The maximum $R^2$ was at $M_f = 80$ (0.4790) for the test data ($R^2$ for the training data was 0.5987 at $M_f = 80$).

**Generalization performance by each group of the field:** The present approach focuses on a specific field and performs optimization concerning each group of the field, so that $R^2$ for the training and test data are obtained for each group. The number of features selected for each group is set to the optimal value $M_f = 80$ determined above, and age is used as the field of interest, dividing it into four groups by quartiles: young, middle-aged A, middle-aged B, and elderly. Table 2 and Figure 2 show the result.

The $R^2$ for the test data for Group 0 (young) is significantly low, but there are two possible reasons. The wide age range of the group (19-38 years) may have accentuated the relative variability of the group, which is more diverse than the other groups. The second possibility may be that the feature design of the original data itself is problematic and needs to capture the younger age group's characteristics adequately. This is also suggested by the low goodness of fit of the group to the training data.

**Table 2:** Coefficients of determination ($R^2$) by age field group

| Group ID | | $R^2$ (training) | $R^2$ (test) |
|---|---|---|---|
| 0 | young (19 - 38 years) | 0.4673 | 0.2761 |
| 1 | middle-aged A (39 - 49 years) | 0.6841 | 0.3198 |
| 2 | middle-aged B (50 - 58 years) | 0.6312 | 0.6062 |
| 3 | elderly (59 - 79 years) | 0.5333 | 0.4021 |

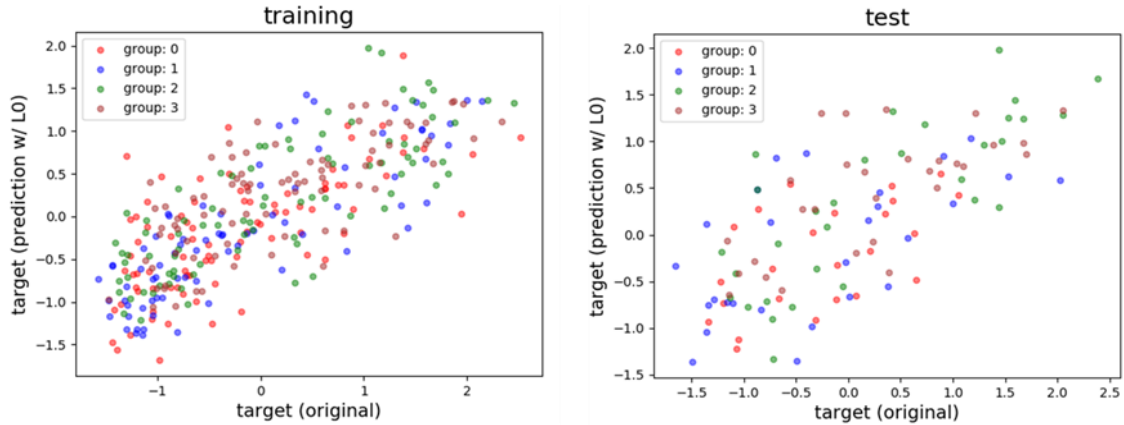

Figure 2: Scatter plot of the target value of the original data (progressive degree of diabetes after one year) and the predicted value by the proposed method (FFM+$L_0$[Ising]), displaying the four groups of the age field separately (left: training data, right: test data). In this figure,

we plotted standardized data values.

Figure 3 shows the relationship between the weight value $(\alpha_s)$ of the expanded feature and the number of groups that selected that feature. In other words, points (on the vertical axis) of 0 indicate that the feature was not selected for any group, corresponding to 540 of the 741 expanded features. One hundred twenty-four features were selected for only one group, meaning features unique to each group. There are 43 (26) features selected for two (three) groups, and eight were selected for all groups regardless of age category.

The six with the most significant absolute value correspond to the first-order weights $(w_l)$ in the eight weights. Note that $\alpha_s > 0$ $(\alpha_s < 0)$ promotes (suppresses) the progression of the disease, and the rightmost point a $(\alpha_s = +0.826)$ corresponds to the category with the highest value in the BMI field. The result indicates that diabetes tends to worsen in the obese group regardless of age category. On the other hand, point b $(\alpha_s = -0.8640)$ and point c $(\alpha_s = -0.4933)$ which inhibit the disease progression regardless of age, correspond to the categories with the lowest or second lowest serum triglycerides, respectively.

Features selected for only one group may reflect the characteristics of the group. Point d $(\alpha_s = -0.8096)$ is Group 0 (young) and corresponds to the female category. These correlations seem to be consistent with the common knowledge assumed from diabetes. On the other hand, we observe a curious correlation in point e. The point represents the correlation of reduced disease progression in the highest LDL (low-density lipoprotein) category in Group 2 (middle-aged B). This way, the present approach can quickly obtain helpful information for exploratory data analysis.

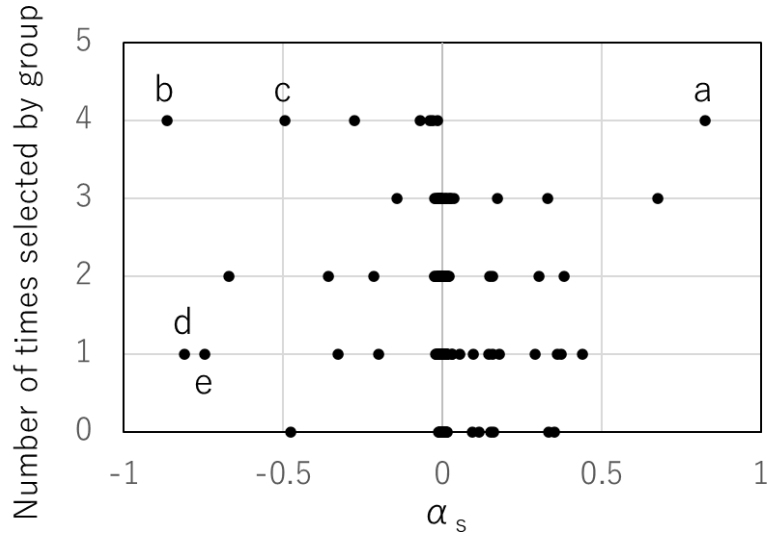

Figure 3: Relationship between the weight $(\alpha_s)$ of an expanded feature and the number of groups selected for that feature.

## Summary

We proposed a feature selection method based on $L_0$ regularization using the Ising model to improve the generalization performance of FFM. In this study, quantitative variables were converted to categorical variables to facilitate the application of the Ising model, and the generalization performance was almost equivalent to that of random forest and elastic net using the original quantitative variables. Therefore, this variable conversion does not cause a considerable loss of information. Another advantage of this method is that selecting appropriate features for each group is possible simultaneously. It makes it possible to indicate for which groups the model under consideration is suitable and for which groups it is not—furthermore, keeping the number of selected features small leads to an intuitive interpretation of the model.

## Acknowledgements

This paper was (partly) based on the results obtained from a project, JPNP16007, commissioned by the New Energy and Industrial Technology Development Organization (NEDO), Japan.